\documentclass{article} 
\usepackage{colm2024_conference}

\usepackage{microtype}
\usepackage{hyperref}
\usepackage{url}
\usepackage{booktabs}
\usepackage{booktabs}
\usepackage{enumitem}
\usepackage{multirow}
\usepackage{array}

\title{Survey of Bias In Text-to-Image Generation: Definition, Evaluation, and Mitigation}

\usepackage{authblk}
\usepackage[symbol]{footmisc}

\author[$\dagger$]{Yixin Wan}
\author[$\dagger$]{Arjun Subramonian}
\author[$\dagger$]{Anaelia Ovalle}
\author[$\dagger$]{Zongyu Lin}
\author[$\dagger$]{Ashima Suvarna}
\author[$\dagger$]{Christina Chance}
\author[$\ddagger$]{Hritik Bansal}
\author[$\ddagger$]{Rebecca Pattichis}
\author[$\dagger$]{Kai-Wei Chang}
\affil[ ]{University of California, Los Angeles}
\affil[$\dagger$]{\texttt{\{elaine1wan, arjunsub, anaelia, linzongy21, asuvarna31, cchance, kwchang\}@cs.ucla.edu}}
\affil[$\ddagger$]{\texttt{\{hbansal, pattichi\}@g.ucla.edu}}

\newcommand{\myparagraph}[1]{\vspace{0.2em}\noindent\textbf{#1}}

\colmfinalcopy
\begin{document}

\maketitle

\begin{abstract}
The recent advancement of large and powerful models with Text-to-Image (T2I) generation abilities---such as OpenAI's DALLE-3 and Google's Gemini---enables users to generate high-quality images from textual prompts.
However, it has become increasingly evident that even simple prompts could cause T2I models to exhibit conspicuous social bias in generated images.
Such bias might lead to both allocational and representational harms in society, further marginalizing minority groups.
Noting this problem, a large body of recent works has been dedicated to investigating different dimensions of bias in T2I systems.
However, an extensive review of these studies is lacking, hindering a systematic understanding of current progress and research gaps.
We present the first extensive survey on bias in T2I generative models.
In this survey, we review prior studies on $3$ dimensions of bias: \textit{Gender},
\textit{Skintone}, and \textit{Geo-Culture}.
Specifically, we discuss how these works \textit{define}, \textit{evaluate}, and \textit{mitigate} different aspects of bias. 
We found that: (1) while gender and skintone biases are widely studied, geo-cultural bias remains under-explored; (2) most works on gender and skintone bias investigated occupational association, while other aspects are less frequently studied; (3) almost all gender bias works overlook non-binary identities in their studies; (4) evaluation datasets and metrics are scattered, with no unified framework for measuring biases; and (5) current mitigation methods fail to resolve biases comprehensively.
Based on current limitations, we point out future research directions that contribute to human-centric definitions, evaluations, and mitigation of biases.
We hope to highlight the importance of studying biases in T2I systems, as well as encourage future efforts to holistically understand and tackle biases, building fair and trustworthy T2I technologies for everyone.
\end{abstract}
\vspace{-1.5em}

\section{Introduction}
\vspace{-0.3em}
Text-To-Image (T2I) models generate accurate images according to textual prompts.
As modern T2I systems such as OpenAI's DALLE-3~\citep{OpenAI_2023} quickly advance in generation quality and prompt-image alignment, many applications in real-world scenarios have been made possible.
AI-generated images are used in political campaigns~\citep{youtubeBeatBiden}, films and TV series~\citep{genai_hollywood, genaifilm}, games~\citep{genai_games}, as well as customized advertisements~\citep{deepagencyModellingAgency}.
Some predict that by 2025, 90\% of internet content will be AI-generated~\citep{genai_content}.
However, as our lives are increasingly infiltrated by AI-created visual content, there is an essential question to consider: \textit{``What does the world depicted by T2I models look like?''}
Prior works have unveiled severe biases in this depiction~\citep{cho2023dalleval,bird2023typology, friedrich2023fair}.
For instance, a version of Stable Diffusion~\citep{rombach2021highresolution} portrayed the world as being run by white masculine CEOs; dark-skinned men are depicted to be committing crimes, while dark-skinned women are delineated to be flipping burgers~\citep{genai_blbg}.
\textbf{T2I models' worldviews are extremely biased, failing to represent the world's diversity of genders, racial groups, and cultures}~\citep{luccioni2023stable}.

Observing the problems with bias in T2I models, we raise a second question: \textit{``Why are such stereotypes and bias concerning?''}
The answer lies in the social risks they could bring.
Previous works warned that stereotypes and bias could cause \textit{representational harms} and \textit{allocational harms} in society~\citep{barocas2017problem, crawford2017trouble, blodgett-etal-2020-language}.
For instance, a study showed that Stable Diffusion depicts over 80\% of ``inmates'' with dark skin~\citep{genai_blbg}, while people of color only constitute less than half of U.S. prison population~\citep{genai_blbg, bopStatisticsInmate}.
Such bias might induce false convictions in real world, if the model is applied to help sketch suspected offenders ~\citep{genai_blbg, pal2023gaussian}.
Along similar lines, ~\citet{wan2024male} discovered that T2I models magnify occupational gender bias in society; ~\citet{bianchi2023easily} found that these models default to depicting Western cultures, whilst under-representing others.
Such biases could result in the reinforcement of dominant cultures, propagation or amplification of social stereotypes, under-representation, and even the erasure of socially marginalized communities~\citep{bianchi2023easily,naik2023social,ungless-etal-2023-stereotypes, solaiman2023evaluating, katirai2023situating, friedrich2024multilingual}.
Furthermore, biases can be propagated by users~\citep{munn2023unmaking} who are unaware of T2I models' underlying issues and trust their output ~\citep{fraser2023diversity}.
Noticing the problem with bias in T2I models, researchers have made efforts to identify different aspects of bias-related risks~\citep{bird2023typology, katirai2023situating}, as well as develop methods for evaluating and mitigating such issues.
However, little has been done to extensively survey bias definitions and methodologies in previous papers.

Towards understanding how prior studies approached bias in T2I models,
we collect $36$ related papers and review their definitions, evaluation methods, and mitigation strategies for biases.
Details on literature collection are in Appendix \ref{appendix:lit-selection}.
\begin{table}[t]
\vspace{-0.8em}
\scriptsize
\begin{center}
\begin{tabular}{p{2cm}>{\raggedright}p{2.75cm}p{8.1cm}}
 \toprule
  \textbf{Category} & \textbf{Conceptualization} & \textbf{Works} \\
  \midrule
   \textbf{Gender Bias} & Default Generation & \citet{naik2023social,zameshina2023diverse,zhang2023inclusive,he2024debiasing,chinchure2023tibet,garcia2023uncurated, bakr2023hrs,esposito2023mitigating}\\
   \cmidrule{2-3}
   & Occupational Association & \citet{cho2023dalleval,bansal2022well,naik2023social,bianchi2023easily,seshadri2023bias,friedrich2023fair,Orgad2023EditingIA,destereotypingTexttoimage,zhang2023inclusive, chinchure2023tibet,shen2023finetuning, luccioni2023stable,fraser2023friendly, vice2023quantifying,li2023fair, zhang2023auditing, wan2024male,mandal2023multimodal,wang-etal-2023-t2iat,lin2023wordlevel,lee2024holistic,sathe2024unified,friedrich2024multilingual}\\
   \cmidrule{2-3}
   & Characteristics and Interests & \citet{naik2023social,bianchi2023easily,li2023fair,fraser2023friendly,zhang2023auditing,friedrich2024multilingual,wang-etal-2023-t2iat,mandal2023multimodal,fraser2023diversity} \\
   \cmidrule{2-3}
   & Stereotypical Objects & \citet{mannering2023analysing,bansal2022well,mandal2023multimodal} \\
   \cmidrule{2-3}
   & Image Quality & \citet{naik2023social,lee2024holistic} \\
   \cmidrule{2-3}
   & Power Dynamics & \citet{wan2024male} \\
   \cmidrule{2-3}
   & NSFW \& Explicit Content & \citet{ungless-etal-2023-stereotypes,hao2024harm} \\
  \midrule 
  \textbf{Skintone Bias} & Default Generation & \citet{naik2023social,zhang2023inclusive,luccioni2023stable,chinchure2023tibet,bakr2023hrs,esposito2023mitigating,he2024debiasing}\\ 
  \cmidrule{2-3}
  & Occupational Association & \citet{bansal2022well,cho2023dalleval,naik2023social,bianchi2023easily,shen2023finetuning, fraser2023friendly,zhang2023inclusive,lee2024holistic} \\
  \cmidrule{2-3}
   & Characteristics and Interests & \citet{naik2023social,fraser2023friendly,wang-etal-2023-t2iat,fraser2023diversity,bianchi2023easily}\\
   \midrule
  \textbf{Geo-Cultural Bias} & Geo-Cultural Norms & \citet{naik2023social,bianchi2023easily,liu2024scoft,basu2023inspecting}\\
  \cmidrule{2-3}
  & Characteristics and Interests & \citet{bianchi2023easily,Jha2024BeyondTS,struppek2023exploiting} \\
  \bottomrule
\end{tabular}
\vspace{-0.8em}
\end{center}
\caption{\label{tab:works}Collection of papers on biases in T2I models in our study, stratified by bias categories and conceptualizations. A large body of works investigated occupational gender biases, whereas only a few study aspects like power dynamics and geo-cultural characteristics.}
\vspace{-1.2em}
\end{table}
We categorize bias definitions along $3$ primarily studied dimensions in previous works: \textbf{Gender Presentation Bias}, \textbf{Skintone Bias}, and \textbf{Geo-Cultural Bias}.
Through extensive literature review on the $3$ dimensions of biases, we observe several limitations and research gaps in current studies:
(1) while as many as $32$ of $36$ papers explored gender bias, only $6$ studied geo-cultural biases; (2) among gender bias studies, $23$ out of $32$ explored occupational biases, while dimensions like image quality and power dynamics are under-studied; (3) only $6$ papers studied bias for non-binary genders, while most works overlook gender minority groups; (4) benchmarks and evaluation methods varied from study to study, indicating a lack of unified frameworks for measuring biases; and (5) current mitigation methods fail to efficiently and effectively resolve biases, with the Gemini controversy being an evident example.
Based on these insights, we identify future steps toward human-centric bias definition, evaluation, and mitigation approaches.
This survey provides a broader perspective on \textit{``what has been done''} and \textit{``what can be done''} for biases in T2I models, lays out solid knowledge foundations and inspirations for future work, and shed light on potential directions of AI governance.

\section{Bias Definitions}
\vspace{-0.8em}
We observe that prior studies, though all focusing on biases in T2I models, provided different definitions of bias \citep{blodgett2021stereotyping}.
It is important to (1) ensure that bias definitions are grounded in social harms~\citep{blodgett-etal-2020-language}, and (2) understand similarities and discrepancies in proposed bias definitions, so as to avoid miscommunication when discussing ``bias''.
Based on our literature survey, we identified that previous explorations on bias in T2I models fall into $3$ major categories: \textbf{Gender Bias}, \textbf{Skintone Bias}, and \textbf{Geo-Cultural Bias}.
In this section, we inductively discuss and summarize how prior works conceptualize these different categories. 
Importantly, the conceptualizations in our taxonomy are not mutually-exclusive.
\vspace{-0.5em}

\subsection{Gender Bias in T2I Models}
T2I models reflect gender stereotypes, such as depicting a ``hairdresser'' with feminine characteristics and ``manager'' with masculine characteristics~\citep{wan2024male}.
Since only gender presentation and roles may be perceived from model-synthesized images, the concept of ``gender'' in these studies refers to \textit{perceived gender presentation and roles}, not gender or sexual identity.
Additionally, most papers that we surveyed defined ``gender'' as a binary concept---only $6$ out of $36$ papers explored bias issues beyond the binary gender.

\myparagraph{Bias in Default Generation} \;
Several prior works defined gender bias as the model's tendency to portray a particular gender when given gender-neutral prompts (e.g. generate ``a person'' or ``a face'')~\citep{naik2023social,zameshina2023diverse,zhang2023inclusive,esposito2023mitigating,he2024debiasing}.
~\citet{chinchure2023tibet} investigated bias in generations for varied prompts from diverse sources.
~\citet{garcia2023uncurated} studied bias in images generated from captions describing real pictures.
~\citet{bakr2023hrs} operationalized this bias as the spurious correlations with gender when the prompt is gender-agnostic.

\myparagraph{Bias in Occupational Association} \;
Most of the reviewed literature on gender bias in T2I models studied occupational gender bias.
A majority of works in this direction defined bias as the model's tendency to over-represent or under-represent a particular gender for an occupation~\citep{cho2023dalleval,bansal2022well,naik2023social,bianchi2023easily,seshadri2023bias,friedrich2023fair,Orgad2023EditingIA,destereotypingTexttoimage,zhang2023inclusive,chinchure2023tibet, shen2023finetuning, luccioni2023stable, fraser2023friendly, vice2023quantifying, li2023fair, zhang2023auditing,lin2023wordlevel,lee2024holistic,sathe2024unified, friedrich2024multilingual,wan2024male}.
~\citet{mandal2023multimodal} and ~\citet{wang-etal-2023-t2iat} operationalized this bias as the embedding distance between a particular gender and stereotypical occupations.

\myparagraph{Bias in Characteristics and Interests} \;
Several previous studies defined bias to be the tendency to generate a particular gender when prompted to depict individuals with certain characteristics or interests like ``intelligent''~\citep{naik2023social,bianchi2023easily,li2023fair,fraser2023friendly,zhang2023auditing,friedrich2024multilingual}.
~\citet{wang-etal-2023-t2iat} operationalized this bias as the embedding proximity between a gender and stereotypical interests, such as ``science'' with men and ``art'' with women.
~\citet{mandal2023multimodal} extended this analysis to interests like ``shopping.''
~\citet{fraser2023diversity} briefly explored biased associations between gender and socioeconomic characteristics. 

\myparagraph{Bias in Stereotypical Objects} \;
~\citet{mannering2023analysing} defined gender bias as a model's propensity to generate gender-stereotypical objects, such as ``ties'' for men and ``handbags'' for women.
~\citet{bansal2022well} assess whether models tend to depict a certain gender when prompted to generate people wearing stereotypical items.
~\citet{mandal2023multimodal} operationalized this biased association in terms of embedding proximity.

\myparagraph{Bias in Image Quality} \;
~\citet{naik2023social} and ~\citet{lee2024holistic} coneptualized gender bias in image quality as the tendency to generate higher-quality images (e.g. with better alignment) for a particular gender.

\myparagraph{Bias in Power Dynamics} \;
~\citet{wan2024male} defined gender bias in power dynamics as the tendency to generate a stereotypical gender for powerful/powerless-indicative prompts, such as men for ``CEO'' or women for ``assistant.''

\myparagraph{NSFW \& Explicit Content} \;
~\citet{ungless-etal-2023-stereotypes} defined bias as the tendency to output 
Not Safe For Work (NSFW) contents for non-cisgender individuals. 
~\citet{hao2024harm} established bias as the disproportionate amplification of input harmful contents for feminine generations.

\subsection{Skintone bias in T2I Models}
T2I models tend to generate social stereotypes related to \textit{perceived} skin tone.
For example, models are shown to reinforce the ``white ideal'' by depicting ``attractive'' individuals as white and ``poor'' individuals as of color~\citep{bianchi2023easily}.

\myparagraph{Bias in Default Generation} \;
Several works conceptualized this aspect of bias as the model's tendency to generate individuals of a certain skintone when skintone was not explicitly specified in the prompt~\citep{naik2023social,zhang2023inclusive,luccioni2023stable,bakr2023hrs,esposito2023mitigating,he2024debiasing}.
~\citet{chinchure2023tibet} studied bias in default generation given varied prompts.

\myparagraph{Bias in Occupational Association} \;
Related studies mostly defined this dimension of bias to be the model's tendency to over- or under-represent skin tone groups when depicting certain occupations~\citep{bansal2022well,cho2023dalleval,naik2023social,bianchi2023easily,shen2023finetuning, fraser2023friendly,zhang2023inclusive,lee2024holistic}.

\myparagraph{Bias in Characteristics and Interests} \;
Most related works defined this bias as the model's tendency to depict a specific skin tone when prompted to generate individuals with certain characteristics, such as ``poor,'' ``pleasant,'' or ``a  criminal''~\citep{naik2023social,fraser2023friendly,wang-etal-2023-t2iat,fraser2023diversity,bianchi2023easily}.

\subsection{Geo-Cultural bias in T2I Models}
T2I models often generate people and artifacts from over-represented cultures or geographical regions---such as the United States~\citep{basu2023inspecting}.

\myparagraph{Bias in Geo-Cultural Norms}
~\citet{naik2023social} and ~\citet{bianchi2023easily} defined bias in cultural norms as the tendency to over-represent specific cultures in the default generation setting while under-representing others.
~\citet{liu2024scoft} and ~\citet{basu2023inspecting} studied the biased norms in depicting non-sensitive words such as clothing and city.

\myparagraph{Bias in Characteristics and Interests}
~\citet{bianchi2023easily} conceptualized this bias as the tendency to depict certain cultures with harmful stereotypes ---such as portraying ``poverty''-indicative images of Africa.
~\citet{Jha2024BeyondTS} further studied biased associations with geo-culturally stereotypical facial features.
~\citet{struppek2023exploiting} extended the exploration to biased associations with scripts of certain languages.

\section{Bias Evaluation}
To understand how previous works measure biases, we review the \textbf{Evaluation Datasets} and \textbf{Evaluation Metrics} in prior studies.
For dataset, we observed a dominance of using different sets of handcrafted prompts, and a lack of unified evaluation benchmarks.
Similarly, we noticed an absence of unified evaluation frameworks and metrics.
In Appendix \ref{appendix:evaluation} Table \ref{tab:eval-works}, we present a table that illustrates the landscape of evaluation metrics in prior works.

\subsection{Evaluation Datasets}

\myparagraph{Manually-Curated Prompts} \;
Several studies constructed and utilized hand-crafted lists of diagnostic prompts with information about targeted bias aspects, such as occupations and characteristics~\citep{hao2024harm, cho2023dalleval,bansal2022well,naik2023social,bianchi2023easily,seshadri2023bias,mandal2023multimodal,wang-etal-2023-t2iat,Orgad2023EditingIA,fraser2023diversity,ungless-etal-2023-stereotypes,destereotypingTexttoimage,zameshina2023diverse,li2023fair, zhang2023inclusive,shen2023finetuning, luccioni2023stable, fraser2023friendly,mannering2023analysing,lin2023wordlevel,basu2023inspecting,esposito2023mitigating,he2024debiasing,wan2024male}.
Additionally, some works created prompts by combining multiple pieces of information, such as gender indicators, human characteristics, and actions~\citep{zhang2023auditing, sathe2024unified,vice2023quantifying}.

\myparagraph{Proposed Datasets}
~\citet{garcia2023uncurated} augmented the Google Conceptual Captions (GCC) dataset~\citep{sharma-etal-2018-conceptual} with age, gender, skintone, and ethnicity annotations.
~\citet{friedrich2024multilingual} collected and proposed the Multilingual Assessment of Gender Bias in Image Generation (MAGBIG) dataset with a variety of prompts in $10$ languages.
~\citet{liu2024scoft} proposed the Cross-Cultural Understanding Benchmark (CCUB) cultural representation dataset for culturally-aware image generation.
~\citet{Jha2024BeyondTS} introduced the ViSAGe (Visual Stereotypes Around the Globe) dataset to assess nationality-based stereotypes in T2I models.
~\citet{lee2024holistic} combined self-curated prompts and data from MS-COCO~\citep{lin2014microsoft} and proposed the Holistic Evaluation of Text-to-Image Models (HEIM) benchmark with bias and fairness subcategories.
~\citet{chinchure2023tibet} combined handcrafted prompts and selected ones from the DiffusionDB database~\citep{wang-etal-2023-diffusiondb}.
~\cite{bakr2023hrs} combined template-based GPT-3.5-generated prompts \cite{ouyang2022training} and DiffusionDB to release a benchmark with 39k prompts (9k for bias evaluation).

\myparagraph{Predefined Datasets}
~\citet{friedrich2023fair} used a subset of LAION-5B for evaluating occupational gender bias.
\citet{vice2023quantifying} utilized $64$ prompts from MS-COCO~\citep{lin2014microsoft}.

\subsection{Evaluation Metrics}
We taxonomized bias evaluation metrics into \textbf{Classification-Based Metrics}---where characteristics are directly inferred---and \textbf{Embedding-Based Metrics}, where characteristics might be subtle and measured in latent space.
We observe that a majority of studies adopted classification-based metrics.
However, ~\citet{zhang2023auditing} pointed out that (1) identities like gender should not be determined only by appearance, and (2) current classifiers fail to represent transgender individuals.
Future studies can explore other alternatives of attribute classifiers to resolve ethical concerns.

\subsubsection{Classification-Based Metrics}
A majority of the literature evaluated bias by classifying demographic characteristics in generated images, such as skintones of depicted individuals.
These works mostly reported bias metrics based on the \textit{level of parity in demographic distribution}, such as the percentage of men and women, or representation of different cultures, in generated images.

\myparagraph{Human Annotation} \;
~\citet{bansal2022well,naik2023social}, ~\citet{wang-etal-2023-t2iat}, ~\citet{fraser2023diversity}, ~\citet{garcia2023uncurated}, ~\citet{fraser2023friendly}, and ~\citet{wan2024male} utilized human-annotated demographics (e.g. gender, skintone, culture) in generated images for evaluation.
~\citet{garcia2023uncurated} reported the percentage difference of unsafe images---as detected by Stable Diffusion v1.4's Safety Checker module---in real images vs. feminine generations from the images' captions.
~\citet{liu2024scoft} asked annotators to rank generated images based on best cultural representation and least stereotypical/offensive.
~\citet{basu2023inspecting} utilized ratings by human annotators to measure geographical representativeness and realism.
~\citet{zhang2023auditing} utilized human-annotated attribute occurrences in image generations.
~\citet{Jha2024BeyondTS} conducted a large-scale annotation task to detect prevalent visual stereotypes in images of 135 identity groups.
~\citet{ungless-etal-2023-stereotypes} conducted human evaluation to classify the quality of images generated for non-cisgender people.

\myparagraph{Classifier-Based Classification}\;
Several works utilized CLIP's zero-shot classification abilities to label the demographics of individuals in generated images~\citep{bansal2022well,cho2023dalleval,seshadri2023bias, Orgad2023EditingIA,zhang2023inclusive,lin2023wordlevel,lee2024holistic,destereotypingTexttoimage}.
A number of other studies used the pre-trained FairFace classifier~\citep{karkkainen2019fairface} to annotate characteristics~\citep{friedrich2023fair,friedrich2024multilingual,naik2023social}.
~\citet{shen2023finetuning} trained their own gender and race classifiers for evaluation.
~\citet{zhang2023inclusive} utilized the Individual Typology Angle (ITA)~\citep{Chardon1991SkinCT} and the method proposed by ~\citet{feng2022towards} to conduct ``unbiased'' skintone classification.
~\citet{cho2023dalleval} used FAN~\citep{bulat2017far}, TRUST~\citep{feng2022towards}, and ITA to map characteristics of individuals in generated images to their closest skintone in the Monk Skin Tone (MST)~\citep{mst_scale} scale.
~\citet{bakr2023hrs} used ArcFace~\citep{deng2019arcface} and RetinaFace~\citep{deng2020retinaface} to detect the faces in the generated images, and then used Dex~\citep{rothe2015dex} to detect facial characteristics related to gender and skintone.
Most of the above studies adopted parity-based metrics~\citep{cho2023dalleval,bansal2022well,seshadri2023bias,Orgad2023EditingIA,lee2024holistic,destereotypingTexttoimage,friedrich2023fair,friedrich2024multilingual,naik2023social,zhang2023inclusive,cho2023dalleval,bakr2023hrs}.
Besides parity measurements, ~\citet{lin2023wordlevel} reported how each word in prompts influences biases in generations.
~\citet{seshadri2023bias} reported percentage differences of feminine images in model generations vs. the training dataset to estimate gender bias amplification.
~\citet{zameshina2023diverse} applied DeepFace~\citep{Taigman2014DeepFace} to detect the gender of individuals in generations, and reported percentage improvement of gender representation after bias mitigation.
~\citet{zhang2023auditing} did not directly classify gender; they trained an attribute classifier to detect elements like ``trousers'', and proposed the Gender Presentation Differences (GEP) Score to measure the overall attribute-wise occurrence differences between images generated from gender-explicit vs. gender-neutral prompts.
~\citet{mannering2023analysing} used You Only Look Once (YOLO) v3~\citep{Redmon2018YOLOv3AI} for object detection, and listed objects in generations for males and females.
~\citet{hao2024harm} employed safety classifiers for text and image to measure the amplification of unsafe/harmful contents in generations compared to inputs. 

\myparagraph{VQA-Based Classification}\;
~\citet{luccioni2023stable} used BLIP~\citep{li2022blip}, a Visual Question Answering (VQA) model to detect the genders of individuals in generated images.
~\citet{cho2023dalleval,esposito2023mitigating,sathe2024unified}, and ~\citet{wan2024male} used BLIP-2~\citep{li2023blip2} for gender classification, but ~\citet{wan2024male} found that BLIP-2 fails in gender classification for complicated images, e.g. with multiple people.
Besides works that adopted parity-based metrics~\citep{luccioni2023stable,cho2023dalleval,esposito2023mitigating,sathe2024unified}, ~\citet{wan2024male} proposed the Paired Stereotype Test (PST) framework to study biases in multi-person generations, and reported the Stereotype Test Scores (STS) to measure the frequency of occurrence of gender stereotypes in generations.
~\citet{esposito2023mitigating} and ~\citet{struppek2023exploiting} used BLIP-2 for skintone and culture classification, respectively.
~\citet{struppek2023exploiting} reported VQA Bias as the percentage increase in culture or skin tone representations when culture-specific characters are added to prompts.
~\citet{vice2023quantifying} used BLIP to extract subjects in model output, and reported distribution bias in generated objects, hallucination level, and the generative miss rate (i.e. percentage of outputs that do not align with prompts).
~\citet{chinchure2023tibet} proposed the Text to Image Bias Evaluation Tool (TIBET), which uses the Concept Association Score (CAS) to measure alignment between MiniGPT-v2~\citep{chen2023minigptv2}-extracted elements in a generated image vs. in generations from perturbed biased prompts.

\myparagraph{Embedding Distance-Based Classification}\;
~\citet{bianchi2023easily, naik2023social}, and ~\citet{basu2023inspecting} labeled characteristics in generated images by comparing their CLIP~\citep{Radford2021LearningTV} embedding distances with a set of other images with pre-defined characteristics (e.g. race, gender).
~\citet{lee2024holistic} compared the similarities of skin pixels to MST
scales to classify skintones.
~\citet{li2023fair} did not specify details, but their visualization illustrated the proximity of images to different genders.
~\citet{he2024debiasing} used FaceNet \cite{schroff2015facenet} to identify faces, and then matched them to descriptions of gender or skintone using text-image CLIP similarity.
While above works mostly used parity-based metrics~\citep{bianchi2023easily,naik2023social,lee2024holistic,li2023fair}, ~\citet{basu2023inspecting} reported Geographical Representativeness (GR), the average realism rating of generations for different countries.
~\citet{bianchi2023easily} reported percentage differences of non-white individuals in generated images vs. real-world statistics to estimate the amplification of societal bias.
~\citet{Jha2024BeyondTS} identify stereotypes in generations by (1) using CLIP to get top 50 captions for images, 
and (2) string-match for stereotypes in captions.
They reported the tendency to generate stereotypical attributes compared with random elements, as well as the offensiveness level.
~\citet{vice2023quantifying} reported distribution bias in (1) generated objects, (2) hallucination level, and (3) generative miss rate, i.e., percentage of outputs that do not align with prompts.

\subsubsection{Embedding-Based Metrics}
\myparagraph{Association with Biased Characteristics} \;
~\citet{wang-etal-2023-t2iat} proposed the T2I Association Test (T2IAT).
They mainly reported differential association, which computes differences in the CLIP~\citep{Radford2021LearningTV}-based proximity of each generation with $2$ other images generated from prompts with opposite sensitive characteristics.
~\citet{mandal2023multimodal} proposed the Multimodal Composite Association Score (MCAS), which is an extension of the Word Embeddings Association Test (WEAT)~\citep{caliskan2017semantics}.
MCAS directly examined the mean cosine distance between the CLIP embedding of a text or a generated image of a particular gender and the embeddings of those of a list of stereotypical items.
~\citet{struppek2023exploiting} replaced Unicode characters in prompts with seemingly identical non-Latin scripts associated with a specific language, and then calculated Relative Bias as the percentage increase in CLIP-embedding proximity with culturally-explicit texts.
~\citet{luccioni2023stable} clustered model outputs into 24 different cluster regions constructed with combined gender and ethnicity characteristics, and reported the shares of cluster assignments to infer attributes.

\myparagraph{Image Quality} \;
~\citet{naik2023social} computed the Frechet Inception Distance (FID)~\citep{heusel2017gans} score between model-generated images for each gender and related web images extracted through an online image search API.
~\citet{lee2024holistic} measured image quality bias by calculating changes in the CLIPScore of images generated from prompts with varied gender or dialect characteristics, compared with the original image-caption pair from MS-COCO~\citep{lin2014microsoft}.

\section{Bias Mitigation}
We categorize bias mitigation strategies into \textbf{Model Weight Refinement}---where models undergo weight adjustments---and \textbf{Inference-Time and Data Approaches}, where model weights remain the same.
Most papers we surveyed adopted prompt-based mitigation approaches, but the method suffers from low robustness and controllability.
In Appendix \ref{appendix:mitigation} Table \ref{tab:mit-works}, we summarize different types of current mitigation approaches.

\subsection{Model Weight Refinement}

\myparagraph{Fine-tuning} \; 
~\citet{struppek2023exploiting} fine-tuned the text encoder to unlearn spurious correlations in T2I generation.
~\citet{esposito2023mitigating} trained a T2I model with generations of another model on a diverse set of prompts, with a combination of ethnicity, gender, age, and profession characteristics from LAION-5B and Stereoset~\citep{nadeem-etal-2021-stereoset}.
~\citet{liu2024scoft} trained T2I models on negative and positive samples of culturally-representative images with a self-contrastive perceptual loss. 
Specifically, they used generations of the pretrained T2I model as negative examples and images from the proposed CCUB dataset as positive ones.

\myparagraph{Parameter-Efficient Fine-Tuning} \;
~\citet{li2023fair}
inserted ``Fair Mapping'', a trainable linear network after the pre-trained text encoder, to map embeddings into a fair space.
During training, they used (1) a fairness loss to diminish associations of bias-sensitive attributes with text embeddings, and (2) a semantic consistency loss to preserve semantic coherence.
~\citet{destereotypingTexttoimage} trained a soft prompt prefix using: (1) a de-stereotyping loss that improves diversity in demographic characteristics of generated images, and (2) a regularization loss to ensure faithful representation of the prompt.
~\citet{shen2023finetuning} performed Low-rank adaptation of large language models (Lora)~\citep{hu2022lora}, fine-tuning with: (1) a distributional alignment loss (DAL) to guide generated characteristics towards a target distribution, and (2) an image semantics-preserving loss.
~\citet{zhang2023inclusive} proposed Inclusive Text-to-Image GENeration (ITI-GEN), which appends a set of learnable prompt tokens to prompts to generate reference images with diverse and inclusive demographic characteristics.

\myparagraph{Model Editing} \;
~\citet{Orgad2023EditingIA} proposed Text-to-Image Model Editing (TIME), which edits the cross-attention layers of Stable Diffusion by aligning the embedding of a gender-neutral input prompt to the embedding of an anti-stereotype gender-specific prompt.

\subsection{Inference-Time and Data Approaches}
\myparagraph{Prompt-Based Mitigation} \;
Several works used ethical intervention prompts, which instruct models with fairness guidelines, to mitigate biases~\citep{bansal2022well,fraser2023diversity,bianchi2023easily,wan2024male}.
However, ~\citet{wan2024male} found that the fairness intervention approach is not fully controllable, resulting in ``overshooting'' biases.
~\citet{naik2023social} and ~\citet{bianchi2023easily} utilized prompts with anti-stereotype demographic characteristics, but found that bias persist despite the instructions.
~\citet{friedrich2024multilingual} experimented with using gender-neutral prompts for languages with grammatical gender or the ``generic masculine''~\citep{SILVEIRA1980165}; however, they found that gender-neutral prompts did not remove gender bias, but rather resulted in a degradation of text-to-image alignment and face generation performance.

\myparagraph{Guided Generation} \;
~\citet{friedrich2023fair} proposed Fair Diffusion, a decoding-time method that extends classifier-free guidance to include textual gender characteristics, towards which the model generation is guided.
Fair Diffusion is able to controllably shift bias in any direction based on human-defined fairness instructions.
~\citet{he2024debiasing} proposed multi-directional guidance paired with iterative distribution alignment, incorporating Kullback–Leibler divergence to guide model output towards a uniform gender distribution.

\myparagraph{Diverse Sampling} \;
~\citet{zameshina2023diverse} proposed finding, in an unsupervised manner, vectors in the latent space of Stable Diffusion that are sufficiently distant from each other to decode so that generated images are diverse.

\myparagraph{Data Augmentation} \;
~\citet{esposito2023mitigating} proposed finetuning T2I models on a diverse set of generated images sourced from different T2I models with varied prompts.

\section{Discussions and Future Directions}
\label{sec:discussions}
Below, we highlight the research gaps present in the study of biases in T2I models, and point out future research directions towards building human-centric approaches for bias definition, evaluation, and mitigation.

\subsection{Human-Centric Approaches for Bias Definitions}
\myparagraph{Clear and Socially-Grounded Conceptualizations of Bias} \;
Some existing literature failed to articulate a clear definition of bias, not clarifying the specific generation task and associated social stereotypes which were being evaluated or mitigated.
For instance, only a few elaborated on their conceptualization of ``gender.'' 
Since only gender presentation and roles may be \textit{perceived} from images, it is important to clarify one's definition, the inherent subjectivity of inferring demographic characteristics, and its implications when discussing research outcomes. 
In addition, works studying skin tone and geo-cultural biases often incorrectly conflate skin tone with race or ethnicity and culture with nationality.
Towards precision and transparency, we encourage researchers to provide details on their conceptualization of bias~\citep{blodgett-etal-2020-language}. 
We further urge researchers to explicitly ground their conceptualizations in concerns about social inequality and power differences~\citep{blodgett2021stereotyping, Ovalle2023Factoring}.

\myparagraph{Extension of Bias Dimensions} \;
Previous studies mostly focused on a limited scope of bias dimensions.
For instance, a large body of works on gender bias explore the unfair associations with occupations.
Meanwhile, only a few research works have explored Geo-Cultural bias in T2I models.
Extending the scope of analysis will provide a more holistic view of the different dimensions of bias (e.g., disability, LGBTQIA+) present in T2I systems.

\subsection{Human-Centric Approaches for Bias Evaluation}
\myparagraph{Community Efforts and Positionality} \;
Socially-grounded evaluations are crucial to AI bias research.
The positionality---subjective experiences, identities, and backgrounds---of researchers and annotators shape their social viewpoints, and might influence their research perspectives ~\citep{santy2023nlpositionality}.
Community efforts seek to understand the influence of technologies on people from different social groups.
Neglecting to examine researcher positionality or solicit human-centric feedback risks missing significant bias considerations, therefore limiting improvements in fairness and inclusivity.
For instance, current works on gender bias depend heavily on binary gender stereotypes.
Consequently, corresponding evaluation strategies are narrowly tailored to address gender binary-specific issues, inadvertently disregarding experiences of individuals beyond the binary, such as transgender and nonbinary communities~\citep{ovalle2023m}.
Researchers should improve community engagement and clarify their positionalities in bias evaluation to improve transparency and trustworthiness.

\myparagraph{Trustworthy Automated Evaluation} \;
Previous works~\citep{luccioni2023stable,cho2023dalleval, esposito2023mitigating,sathe2024unified,wan2024male} explored the use of VQA models for detecting gendered characteristics of individuals in generations.
However, ~\citet{wan2024male} pointed out that current VQA models like BLIP-2 fail to correctly identify gender characteristics in complicated images (e.g. with two people), which has become a bottleneck in advancing automated evaluation.
Furthermore, VQA models might carry underlying biases~\citep{ruggeri-nozza-2023-multi}.
Building stronger and more reliable automated evaluation methods will allow for scalable approaches to evaluating biases in T2I models.

\subsection{Human-Centric Approaches for Bias Mitigation}

\myparagraph{Diverse and Inclusive Mitigation} \;
Previous works proposed bias mitigation techniques based on their conceptualizations of biases.
However, very few considered user preferences when designing solutions for biases \cite{mehrabi2023resolving}; ~\citet{ungless-etal-2023-stereotypes} was among the few to study how non-binary users prefer to be represented in T2I model generations.
We highlight that \textbf{diverse outputs do not imply ``inclusion''}---which refers to the sense of belonging and representation among users~\cite {mitchell2020diversity}---and encourage future researchers to consider this point.

\myparagraph{Robust, Controllable, and Resource-Friendly Mitigation} \;
Current mitigation methods mostly adopt training and prompt-based approaches.
However, training-based approaches require data and resources, and could result in catastrophic forgetting~\citep{luo2023empirical} or degraded generation quality~\citep{yang2022unified}.
Furthermore, prompt-based approaches suffer from a lack of robustness and controllability~\citep{friedrich2024multilingual, wan2024male}, resulting in unstable behaviors such as failing to follow instructions and ``over''-mitigating.
Therefore, designing robust, controllable, and resource-friendly mitigation methods is essential for easy and safe T2I applications.

\myparagraph{Beyond One-size-fits-all Mitigation Approaches} One-size-fits-all bias mitigation for T2I models may be at odds with known challenges, such as their tendency to hallucinate, or produce factual content.
For instance, focusing on generating content with equitable distributions across sensitive characteristics may result in inaccurate and offensive image generations, like in the Gemini incident~\citep{timeEthicalIsnt}.
Among surveyed works, \citet{ungless-etal-2023-stereotypes} discussed how prompt blocking---such as curated system prompts, prompt blacklists, or post-prompt moderation---may contribute to the erasure of non-cisnormative identities.

One potential solution is aligning generations with diverse human and community preferences ~\citep{xu2024imagereward,kirstain2024pick,wu2023human}.
For instance, in language modeling, \cite{bai2022constitutional,bai2022training,touvron2023llama} trained safety reward models can guide model generations towards harmless outputs (e.g., refraining to suggest ways to steal the neighbor's Wi-Fi password).
However, most prior studies on alignment for T2I models \citep{lee2023aligning,fan2024reinforcement,wallace2023diffusion,sun2023dreamsync} used it to enhance the image quality and image-prompt matching;
limited works have explored alignment as a mitigation strategy for societal biases in T2I models.
However, while we encourage future explorations beyond one-size-fits-all approaches, researchers should critically examine how AI biases intersect with methodological limitations.
As such, proxy objectives, reward hacking, and overoptimization are fundamental challenges to navigate when aligning to human preferences \citep{casper2023open,bansal2023peering}.

\myparagraph{Adaptive and Lifelong Mitigation} \;
Conceptualizations of bias, bias dimensions, and bias-related stereotypes  change over time.
Only mitigating currently-identified biases in models is not sufficient, as new problems might continue to appear in a model's life cycle.
Instead of static bias mitigation methods, we encourage future studies to explore dynamic, adaptive, and lifelong mitigation strategies that evolve based on continuous feedback and the changing landscape of societal norms and values, as well as community needs.
For example, involving real-time monitoring of model outputs and automatic adjustment of model parameters might help with addressing emerging or varied conceptualizations of bias.

\section{Conclusion}
In this paper, we present the first survey on biases in T2I models.
We thoroughly reviewed and summarized the definitions, evaluation metrics, and mitigation methods in $36$ previous papers on $3$ dimensions of biases: \textbf{Gender Presentation}, \textbf{Skintone}, and \textbf{Geo-Cultural}.
During our surveying of the papers, we made several meaningful observations: (1) most works focused on studying biases in gender and skintones, whereas very few investigated geo-cultural biases; (2) a majority of gender and skintone bias research explored biased associations with occupations, but few studied aspects like biases in power dynamics and explicit content generation; (3) most works on gender bias do not consider how non-binary individuals are represented; (4) no unified framework for bias evaluation has been consolidated, and bias metrics vary from study to study; and (5) current mitigation methods fail to come up with holistic and effective solutions for biases.
Based on our insights, we highlight future research directions in human-centric approaches for bias definitions, evaluation, and mitigation.
We hope this work offers a helpful overview of the landscape of research on biases in T2I models for researchers and policymakers alike, as well as sheds light on potential paths toward building and regulating trustworthy T2I systems.


\section{Ethics Statement}
This study collects and reviews previous works on biases in T2I models.
Below, we identify a number of important ethical considerations in this paper, as well as in the papers that we survey.

\myparagraph{Inferring Personal Identities From Images} \;
Personal identities in this paper and in related works---such as gender, skintone, and geo-culture---are based on their visual presentation in model-generated images.
For instance, prior works often used identity-indicative characteristics as a proxy of the identity itself: using gender characteristics as a proxy for gender, skintone as a proxy for race, and culturally-unique elements as a proxy for culture. This is due to synthesized individuals in generated images not having internal identities.
In particular, we note that: (1) ``gender'' and ``culture'' in prior works indicate visual presentations of gender and cultural traits, (2) while previous studies sometimes use ``skintone,'' ``race,'' and ``ethnicity'' interchangeably, ``race'' and ``ethnicity'' can never be inferred from images, and (3) some papers incorrectly conflate ``culture'' and ``nationality,'' but many countries comprise a multitude of cultures, albeit with some cultures dominant \citep{Qadri2023Regimes}. Therefore, to ensure clarity of definitions and avoid misconceptions, throughout our survey, we utilized ``gender presentation,'' ``skintone,'' and ``geo-cultural'' to represent dimensions of bias based on visual presentation.

\myparagraph{Classification Biases} \;
We underline the limitations of classification strategies in prior works.
When classifying the identities of individuals in generated images, previous studies adopted methods such as human annotation, model-based classifiers, and embedding distance-based classification.
However, both human annotators~\citep{pandey2019modeling, sap-etal-2022-annotators, biester-etal-2022-analyzing} and automated methods~\citep{6327355, Abdurrahim2018ReviewOT, muthukumar2018understanding, Buolamwini2018GenderSI, mcduff2019characterizing, wu2020gender, krishnan2020understanding, ramachandran2022deep, KRISHNAN2023104793, garcia2023uncurated} have been shown to possess bias.
Therefore, using these classification tools will inevitably risk propagating their bias in T2I bias evaluation results.
We stress that researchers be aware of this limitation, and transparently and critically discuss their approaches to classification.

\myparagraph{AI For Justice-Centered Applications} \;
The definitions and methodologies in this survey should not support evaluations and mitigation for bias in unjust applications of generative AI.
We emphasize that approaches for measuring or resolving biases should not be exploited to ethics-wash applications of generative AI that enable surveillance, harming or suppressing workers~\citep{Jiang2023Art}, weaponization, etc.



\appendix
\section{Literature Selection Methodology}
\label{appendix:lit-selection}
To select the papers for our analysis, we began with an examination of the most-cited papers on the topic of ``bias'' and ``fairness'' for Text-to-Image (T2I) models, as indexed by Google Scholar and arXiv.
As the field is relatively new with a large body of very recent papers, we include both peer-reviewed works and preprints.
This initial step provided a foundation from which we explored further, delving into works cited by and that cited these papers.
This approach was designed to ensure an extensive and informed review of the field, capturing a wide range of perspectives and methodologies.
While we endeavored to be thorough in the literature selection process, we acknowledge the possibility that a small portion of relevant papers may have been inadvertently overlooked.
We recognize the inherent limitations in any literature review or collection process.

\section{Summarization of Evaluation Metrics}
\label{appendix:evaluation}
Table \ref{tab:eval-works} summarizes and stratifies evaluation methods in previous works.
$18$ out of the total of $36$ works that we studied adopted classifier-based approaches to identify demographic characteristics in generated images.
Only a few explored embedding-based measurements such as generation diversity and image quality.

\begin{table}[hb]
\scriptsize
\begin{center}
\begin{tabular}{p{2cm}p{3cm}p{8.1cm}}
 \toprule
  \textbf{Metric Type} & \textbf{Category} & \textbf{Works} \\
  \midrule
   \textbf{Classification-Based Metrics} & Human Annotation & \citet{bansal2022well,naik2023social, wang-etal-2023-t2iat, fraser2023diversity, garcia2023uncurated, fraser2023friendly, basu2023inspecting, zhang2023auditing, ungless-etal-2023-stereotypes, wan2024male, liu2024scoft, Jha2024BeyondTS} \\
   \cmidrule{2-3}
   & Classifier-Based & \citet{bansal2022well,cho2023dalleval,seshadri2023bias, bakr2023hrs, zhang2023inclusive, Orgad2023EditingIA, zameshina2023diverse, shen2023finetuning, naik2023social, friedrich2023fair, lin2023wordlevel,zhang2023auditing, mannering2023analysing,lee2024holistic, destereotypingTexttoimage, friedrich2024multilingual, hao2024harm} \\
   \cmidrule{2-3}
   & VQA-Based & \citet{esposito2023mitigating, luccioni2023stable, cho2023dalleval, vice2023quantifying, struppek2023exploiting, chinchure2023tibet,sathe2024unified, wan2024male} \\
   \cmidrule{2-3}
   & Embedding Distance-Based & \citet{bianchi2023easily, vice2023quantifying, naik2023social, li2023fair, basu2023inspecting, hao2024harm, Jha2024BeyondTS,  lee2024holistic,  he2024debiasing} \\
  \midrule 
  \textbf{Embedding-Based Metrics} & Association with Bias Characteristics & \citet{wang-etal-2023-t2iat,  mandal2023multimodal, struppek2023exploiting, luccioni2023stable} \\ 
  \cmidrule{2-3}
   & Image Quality & \citet{naik2023social, lee2024holistic} \\
  \bottomrule
\end{tabular}
\vspace{-0.8em}
\end{center}
\caption{\label{tab:eval-works} Literatures on bias evaluation metrics for T2I models, stratified by metric types and categories. While many studies employed classification-based metrics, only a few used embedding-based methods.}
\vspace{-1.5em}
\end{table}

\section{Summarization of Mitigation Metrics}
\label{appendix:mitigation}
Table \ref{tab:mit-works} summarizes different types of mitigation approaches in previous works.
While not many have explored bias mitigation in T2I models, most works in this direction explored Parameter-Efficient Finetuning and prompt-based methods.
Other approaches like model editing, inference-time guidance and sampling, and data augmentation remain under-explored.

\begin{table}[htb]
\scriptsize
\begin{center}
\begin{tabular}{p{2.8cm}p{2.75cm}p{7.5cm}}
 \toprule
  \textbf{Category} & \textbf{Conceptualization} & \textbf{Works} \\
  \midrule
   \textbf{Model Weight Refinement} & Finetuning & \citet{struppek2023exploiting, esposito2023mitigating, liu2024scoft} \\
   \cmidrule{2-3}
   & Parameter-Efficient Finetuning & \citet{li2023fair, destereotypingTexttoimage, shen2023finetuning,  zhang2023inclusive} \\
   \cmidrule{2-3}
   & Model Editing & \citet{Orgad2023EditingIA} \\
  \midrule 
  \textbf{Inference-Time and Data Approaches} & Prompt-Based Mitigation & \citet{bansal2022well,fraser2023diversity, bianchi2023easily, naik2023social, friedrich2024multilingual,wan2024male} \\ 
  \cmidrule{2-3}
  & Guided Generation & \citet{friedrich2023fair, he2024debiasing} \\
  \cmidrule{2-3}
   & Diverse Sampling & \citet{zameshina2023diverse}  \\
   \cmidrule{2-3}
   & Data Augmentation & \citet{esposito2023mitigating} \\
  \bottomrule
\end{tabular}
\vspace{-0.8em}
\end{center}
\caption{\label{tab:mit-works} Literatures on bias mitigation approaches for T2I models that we reviewed, stratified by mitigation types and methods. Most prior works employ Parameter-Efficiant Finetuning and Prompt-Based Mitigation, whereas methods like data augmentation and model editing remain under-explored.}
\vspace{-1.5em}
\end{table}

\end{document}